\documentclass[letterpaper]{sig-alternate-2013}

\newfont{\mycrnotice}{ptmr8t at 7pt}
\newfont{\myconfname}{ptmri8t at 7pt}
\let\crnotice\mycrnotice%
\let\confname\myconfname%

\usepackage{graphicx}
\usepackage{amsmath}
\usepackage{amssymb}
\usepackage{algorithmic}
\usepackage{algorithm}
\usepackage{subfigure}

\newcommand{\bfx}{{\textbf{x}}}
\newcommand{\bfu}{{\textbf{u}}}
\newcommand{\bfv}{{\textbf{v}}}
\newcommand{\bff}{{\textbf{f}}}
\newcommand{\bfpi}{{\boldsymbol{\pi}}}

\begin{document}

\title{Cross-Domain Sparse Coding}

\numberofauthors{1}
\author{
\alignauthor
Jim Jing-Yan Wang$^{1,2}$\\
\affaddr{$^1$
University at Buffalo, The State University of New York,
Buffalo, NY 14203, USA}\\
\affaddr{$^2$
National Key Laboratory for Novel Software Technology, Nanjing University, Nanjing 210023, China
}\\
\email{jimjywang@gmail.com}
}

\maketitle

\begin{abstract}
Sparse coding has shown its power as an effective data representation
method.
However, up to now, all the
sparse coding approaches are limited within the single domain learning problem.
In this paper, we extend the sparse coding to cross domain learning problem,
which tries to learn from a source domain to a target domain with significant different distribution.
We impose the Maximum Mean Discrepancy (MMD) criterion
to reduce the cross-domain distribution difference
of  sparse codes,
and also  regularize the sparse codes by the
class labels of the samples from both domains to increase the discriminative ability.
The encouraging experiment results of the
proposed cross-domain sparse coding algorithm
on
two challenging tasks --- image classification of
photograph and oil painting domains, and
multiple user spam detection ---
show
the advantage of the proposed method over other cross-domain
data representation methods.
\end{abstract}


%\terms{Algorithm}


\section{Introduction}

Traditional machine learning methods usually
assume that
there are sufficient training samples to train the classifier.
However,  in many real-world applications, the number of labeled samples are always limited,
making the learned classifier not robust enough.
Recently,  cross-domain learning
has been proposed to solve this problem \cite{Blitzer2006}, by
borrowing labeled samples from a so called ``source domain"
for the learning problem of the ``target domain" in hand.
The samples from these two domains have different distributions but are related, and share
the same class label and feature space.
Two types of domain transfer learning methods have been studied:
\textbf{classifier transfer} method which
learns a classifier for the target domain by the target domain samples with help of
the source domain samples,
while \textbf{cross domain data representation}
tries to map
all the samples from both source and target domains to a data representation space with a common distribution
across domains,
which could be used to train a single domain classifier for the
target domain \cite{Blitzer2006,DaumeIII2007,Pan2009,Pan2011}.
In this paper, we focus on the cross domain representation problem.
Some works have been done in this field by various data representation methods.
For example,
Blitzer et al. \cite{Blitzer2006} proposed
the  structural correspondence learning (SCL)
algorithm to induce correspondence among features from the
source and target domains,
Daume III
\cite{DaumeIII2007} proposed
the feature replication (FR) method to augment
features for cross-domain learning.
Pan et al.
\cite{Pan2009}
proposed transfer component analysis (TCA) which learns transfer components across domains
via Maximum
Mean Discrepancy (MMD) \cite{MMD2006}, and
extended it to semi-supervised  TCA (SSTCA) \cite{Pan2011}.

Recently, sparse coding has attracted many attention as an effective data representation method,
which represent a data sample as
the sparse linear combination of some codewords in a codebook \cite{Lee2007}.
Most of the sparse coding algorithms are unsupervised, due to the
small number of labeled samples.
Some semi-supervised sparse coding methods are proposed  to
utilize the labeled samples and significant performance improvement has been reported \cite{Yang2012}.
In this case,
it would be very interesting to investigate the use of
cross-domain representation to provide
more available labeled samples from the source domain.
To our knowledge, no work has been done using the sparse coding method to solve the
cross-domain problem
To fill in this gap, in this paper, we propose a novel
cross-domain sparse coding method
to combine the advantages of both sparse coding and cross-domain learning.
To this end, we will try to learn a common codebook for the
sparse coding of the samples from both the source and target domains.
To utilize the class labels, a semi-supervised regularization will also be introduced to
the sparse codes.
Moreover, to reduce the mismatch between the distributions of the sparse codes of the
source and target samples, we adapt the MMD rule to sparse codes.

The remaining of this paper is organized as follows:
In Section \ref{sec:CroDomSc}, we will
introduce the formulations of the proposed Cross-Domain
Sparse coding (CroDomSc), and
its implementations.
Section \ref{sec:Exp} reports experimental results, and
Section \ref{sec:Concl} concludes the paper.

\vfill\eject

\section{Cross-Domain Sparse Coding}
\label{sec:CroDomSc}

In this section, we will introduce the proposed
CroDomSc method.

\subsection{Objective function}

We denote the training dataset  with $N$ samples as
$\mathcal{D}=\{\bfx_i\}_{i=1}^{N}\in \mathbb{R}^D$, where
$N$ is the number of data samples,
$\bfx_i$ is the feature vector of the $i$-th sample,
and $D$ is the feature dimensionality.
It is also organized as a  matrix $X=[\bfx_1, \cdots,\bfx_N]\in \mathbb{R}^{D\times N}$.
The training set is composed of the source domain set $\mathcal{D}^S$ and target domain set $\mathcal{D}^T$,
i.e., $\mathcal{D}=\mathcal{D}^S \bigcup \mathcal{D}^T$.
We also denote $N^D$ and $N^T$ as the number of samples in source and target domain set separatively.
All the samples from the source domain set $\mathcal{D}^S$ are labeled, while only a few samples from the
target domain $\mathcal{D}^T$ are labeled.
For each labeled sample $\bfx_i$, we denote its class label as $y_i\in \mathcal{C}$,
where $\mathcal{C}$ is the class label space.
To construct the objective function, we consider the following three
problems:

\begin{description}
\item[Sparse Coding Problem]
Given a sample $\bfx_i\in  \mathcal{D}$ and a codebook matrix $U = [u_1, \cdots, u_K]\in \mathbb{R}^{D\times K}$,
where the $k$-th column is the $k$-th codeword and $K$ is the number of codewords in the codebook,
sparse coding  tries to reconstruct $x$ by the
linear reconstruction of the codewords in the codebook as $\bfx_i \approx \sum_{k=1}^K v_{ki} \bfu_k=U \bfv_i$,
where
$\bfv_i=[v_{1i},\cdots, v_{Ki}]^\top \in \mathbb{R}^{K}$
is the reconstruction coefficient vector for $\bfx_i$, which should be as sparse as possible,
thus $\bfv_i$ is also called sparse code.
The problem of sparse coding can be formulated as
follows:
\begin{equation}
\label{equ:Sc}
\begin{aligned}
\underset{U, V}{min}
&
\sum_{i:\bfx_i\in \mathcal{D}}
\left(\|\bfx_i - U \bfv_i\|^2_2 + \alpha \|\bfv_i\|_1\right)
\\
&
=
\|X - U V\|^2_2 + \alpha \sum_{i:\bfx_i\in \mathcal{D}} \|\bfv_i\|_1
\\
s.t.&\|\bfu_k\|\leq c,~k=1,\cdots,K
\end{aligned}
\end{equation}
where $V=[\bfv_1, \cdots, \bfv_N]\in \mathbb{R}^{K \times N}$ is the sparse code matrix,
with its $i$-th collum the sparse code of $i$-th sample.

\item[Semi-Supervised Sparse Coding Regularization]
In the sparse code space, the intra-class variance should be minimized while the inter-class variance
should be maximized
for all the samples labeled, from both target and source domains.
We first define the semi-supervised regularization matrix as
$W=[W_{ij}]\in \{+1,-1,0\} ^{N\times N}$, where
\begin{equation}
\begin{aligned}
W_{ij}=
\left\{\begin{matrix}
+1, & if~y_i=y_j,\\
-1, &if~y_i\neq y_j,\\
0,& if~y_i~or~y_j~is~unkown.
\end{matrix}\right.
\end{aligned}
\end{equation}
We define the degree of $\bfx_i$ as $d_i=\sum_{j:\bfx_j\in \mathcal{D}}W_{ij}$,
$D=diag(d_1,\cdots,d_N)$, and $L=D-W$ as the
is the Laplacian matrix.
Then we formulate the semi-supervised regularization problem as the following problem:
\begin{equation}
\label{equ:Semisupervised}
\begin{aligned}
\underset{V}{min}
&\sum_{i,j:\bfx_i,\bfx_j\in \mathcal{D}}
\|\bfv_i - \bfv_j\|^2_2 W_{ij}\\
&=tr[V (D-W) V^\top]=tr(V L V^\top)
\end{aligned}
\end{equation}
In this way,
the $l_2$ norm distance between sparse codes of intra-class pair ($W_{ij}=1$) will be minimized,
while inter-class pair ($W_{ij}=-1$) maximized.

\item[Reducing Mismatch of Sparse Code Distribution]
To reduce the mismatch of the distributions of the source domain and target domain
in the sparse code space,
we adopt the
MMD \cite{MMD2006}
as a criterion,
which is based on the minimization of the distance between the
means of codes from two domains.
The problem of reducing the mismatch of the  sparse code distribution between source and target domains
could be formatted as follows,
\begin{equation}
\label{equ:Mismatch}
\begin{aligned}
\underset{V}{min}
&\begin{Vmatrix}
\frac{1}{N_S}\sum_{i:\bfx_i\in \mathcal{D}^S} \bfv_i
-  \frac{1}{N_T}\sum_{j:\bfx_j\in \mathcal{D}^T} \bfv_j
\end{Vmatrix}^2\\
&=\|V\bfpi\|^2_2
=Tr[V\bfpi \bfpi^\top V^\top]=Tr(V \Pi V^\top)
\end{aligned}
\end{equation}
where $\bfpi=[\pi_1,\cdots,\pi_N]^\top \in \mathbb{R}^{N}$ with $\pi_i$ the domain indicator of $i$-th sample defined as
\begin{equation}
\begin{aligned}
\pi_i=
\left\{\begin{matrix}
\frac{1}{N_S}, & \bfx_i\in \mathcal{D}^S,\\
-\frac{1}{N_T}, & \bfx_i\in \mathcal{D}^T.\\
\end{matrix}\right.
\end{aligned}
\end{equation}
and
$\Pi=\bfpi \bfpi^\top$.

\end{description}

By summarizing the formulations in (\ref{equ:Sc}), (\ref{equ:Semisupervised})
and (\ref{equ:Mismatch}),
the CroDomSc problem is modeled as the following optimization problem:
\begin{equation}
\label{equ:Objective}
\begin{aligned}
\underset{U, V}{min}
&\|X-UV\|^2_2+
\beta Tr[V L V^\top]+
\gamma Tr[V \Pi V^\top]
+\alpha \sum_{i:\bfx_i\in \mathcal{D}} \|\bfv_i\|_1
\\
&=\|X-UV\|^2_2+
Tr[V E V^\top]
+\alpha \sum_{i:\bfx_i\in \mathcal{D}} \|\bfv_i\|_1
\\
s.t.&\|\bfu_k\|\leq c,~k=1,\cdots,K
\end{aligned}
\end{equation}
where $E=(\beta L+ \gamma \Pi)$.

\subsection{Optimization}

Since direct optimization of (\ref{equ:Objective}) is difficult,
an iterative, two-step strategy is used to  optimize
the codebook
$U$ and sparse codes $V$ alternately while fixing the other one.

\subsubsection{On optimizing $V$ by fixing $U$}

By fixing the codebook $U$,
the optimization
problem (\ref{equ:Objective}) is reduced to
\begin{equation}
\label{equ:Object_v}
\begin{aligned}
\underset{V}{min}
&\|X-UV\|^2_2+
Tr[V E V^\top]
+\alpha \sum_{i:\bfx_i\in \mathcal{D}} \|\bfv_i\|_1
\end{aligned}
\end{equation}

Since the reconstruction error term can be rewritten as
$\|X-UV\|^2_2
=\sum_{i:\bfx_i\in \mathcal{D}}
\|\bfx_i - U \bfv_i\|^2_2$,
and the sparse code regularization term could be
rewritten as
$Tr[V E V^\top]$~
$=\sum_{i,j:\bfx_i,\bfx_j\in \mathcal{D}}
E_{ij}  \bfv_i^\top \bfv_j$,
(\ref{equ:Object_v}) could be rewritten as:
\begin{equation}
\begin{aligned}
\underset{V}{min}
&\sum_{i:\bfx_i\in \mathcal{D}} \|\bfx_i - U \bfv_i\|^2_2 +
\sum_{i,j:\bfx_i,\bfx_j\in \mathcal{D}}
E_{ij}  \bfv_i^\top \bfv_j
+\alpha \sum_{i:\bfx_i\in \mathcal{D}} \|\bfv_i\|_1
\end{aligned}
\end{equation}
When updating $\bfv_i$ for any $\bfx_i\in \mathcal{D}$, the other codes
$\bfv_j(j\neq i)$ for $\bfx_j\in \mathcal{D}, j\neq i$
are fixed.
Thus, we get the following optimization problem:
\begin{equation}
\label{equ:ScV}
\begin{aligned}
\underset{{\bfv_i}}{min} %_{\bfx_i\in \mathcal{D}^S}
&\|\bfx_i-U\bfv_i\|^2_2+
E_{ii}  \bfv_i^\top \bfv_i +
\bfv_i^\top \bff_i
+\alpha  \|\bfv_i\|_1
\end{aligned}
\end{equation}
with $\bff_i=2 \sum_{j:\bfx_j \in \mathcal{D}, j\neq i} E_{ii}  \bfv_j$.
The
objective function in (\ref{equ:ScV})
could be optimized efficiently by the modified feature-sign search algorithm proposed in \cite{GraphSc2011}.

\subsubsection{On optimizing $U$ by fixing $V$}

By fixing the sparse codes $V$ and removing irrelevant terms, the optimization problem
(\ref{equ:Objective}) is reduced to
\begin{equation}
\label{eqe:Object_U}
\begin{aligned}
\underset{U}{min}
&\|X-UV\|^2_2\\
s.t.&\|\bfu_k\|^2_2\leq c,~k=1,\cdots,K
\end{aligned}
\end{equation}
The problem
is a least square problem with quadratic constraints, and it can be solved
in the same way as \cite{Lee2007}.

\subsection{Algorithm}

The proposed \textbf{Cross Domain Sparse coding} algorithm, named as \textbf{CroDomSc}, is summarized in Algorithm \ref{alg:CroDomSs}.
We have applied the original sparse coding methods to
the samples from both the source and target domains for initialization.

\begin{algorithm}[h!]
\caption{CroDom-Ss Algorithm}
\label{alg:CroDomSs}
\begin{algorithmic}
\STATE \textbf{INPUT}:
Training sample set from both source and target sets $\mathcal{D}=\mathcal{D}^S \bigcup \mathcal{D}^T$;

\STATE Initialize the codebooks ${U}^0$ and sparse codes $V^0$
for samples in $\mathcal{D}$ by using single domain sparse coding.

\FOR{$t=1,\cdots,T$}

\FOR{$i: \bfx_i\in \mathcal{D}$}

\STATE Update the sparse code $\bfv_i^t$ for $\bfx_i$ by fixing
$U^{t-1}$ and other sparse codes $\bfv_j^{t-1}$ for $\bfx_j\in \mathcal{D}, j\neq i$ by solving
(\ref{equ:ScV}).

\ENDFOR

\STATE Update the codebook ${U}^t$ by fixing the sparse code matrix $V^t$
by solving (\ref{eqe:Object_U}).

\ENDFOR

\STATE \textbf{OUTPUT}: ${U}^T$ and $V^T$.

\end{algorithmic}
\end{algorithm}

When a test sample from target domain comes, we simply solve problem (\ref{equ:ScV}) to obtain its
sparse code.

\section{Experiments}
\label{sec:Exp}

In the experiments, we experimentally evaluate the proposed cross domain data representation
method, CroDomSc.

\subsection{Experiment I: Cross-Domain Image Classification}

In the first experiment, we considered the problem of cross domain image classification
of
the photographs and the oil paintings, which are treated as two different domains.

\subsubsection{Dataset and Setup}

We collected an image database of both photographs and oil paintings.
The database contains totally 2,000 images of 20 semantical classes.
There are 100 images in each class, and 50 of them are photographs, and
the remaining 50 ones are oil paintings.
We extracted and concatenated the
color, texture, shape and bag-of-words histogram features as
 visual feature vector from each image.

To conduct the experiment, we use photograph domain and oil painting domain
as source domain and target domain in turns.
For each target domain, we randomly split it into training subset (600 images) and
test subset (400 images), while 200 images from the training subset are randomly selected
as label samples and all the source domain samples are labeled.
The random splits are repeated   for 10 times.
We first perform the CroDomSc to the training set and use the
sparse codes learned to train a semi-supervised SVM classifier \cite{Geng2012}.
Then the test samples will also be represented as sparse code and classified using the
learned SVM.

\subsubsection{Results}

\begin{figure}[htbp!]
\centering
\subfigure[Photograph as target domain]{
\includegraphics[width=0.5\textwidth]{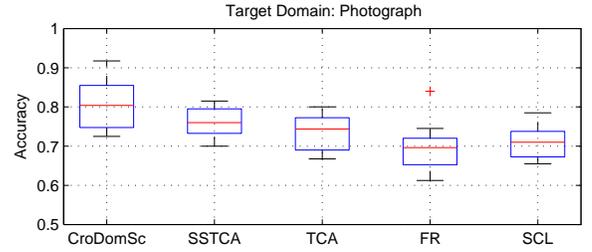}}
\subfigure[Oil painting as target domain]{
\includegraphics[width=0.5\textwidth]{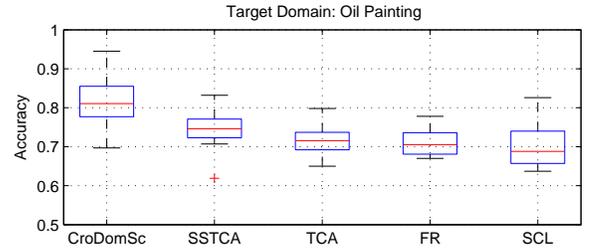}}
\caption{The boxplot of classification accuracies of 10 splits
of CroDomSc and compared methods.}
%, using \textbf{photograph} set as target domain.}
\label{fig:FigPho}
\end{figure}

We compare our CroDomSc against
several cross-Domain data representation methods: SSTCA \cite{Pan2011}, TCA \cite{Pan2009},
FR \cite{DaumeIII2007} and SCL \cite{Blitzer2006}.
The boxplots of the  classification accuracies of the 10 splits
using photograph and oil painting as target domains
are reported in Figure \ref{fig:FigPho}.
From Figure \ref{fig:FigPho} we can see that the proposed CroDomSc outperforms
the other four competing methods
for both photograph and oil painting domains.
It's also interesting to notice that the classification  of  the FR and SCI methods
are poor, at around 0.7.
SSTCA and TCA seems better than FR and SC but are still  not competitive to CroDomSc.

\subsection{Experiment II: Multiple User Spam Email Detection}

In the second experiment, we will evaluate the proposed cross-domain data representation
method for the multiple user based spam email detection.

\subsubsection{Dataset and Setup}

A email dataset with 15 inboxes from 15 different users is used in this experiment \cite{SpamECML2006}.
There are 400 email samples in each inbox, and  half of them are
spam and the other half non-spam.
Due to the significant differences of the email source among different users,
the email set of different users could be treated as different domains.

To conduct the experiment, we
randomly select two users' inboxes as
source and target domains. The target domain will further be  split into test set (100 emails)
and training set (300 emails, 100 of which labeled, and 200 unlabeled).
%In the trained set, 100 emails will be randomly selected as labeled samples,
%while the remaining 200 ones as unlabeled.
The source domain emails are all labeled.
%For the email samples in the source domain, they are all used as labeled samples.
The word occurrence frequency histogram is extracted from each email as original feature vector.
The CroDomSc algorithm was performed to learn the sparse code of both source and target domain samples,
which were used to train the semi-supervised classifier.
The target domain test samples were also represented as sparse codes, which were classified using the
learned classifier.
This selection will be repeated for 40 times to reduce the bias of each selection.

\subsubsection{Results}

\begin{figure}[htbp!]
\centering
\includegraphics[width=0.5\textwidth]{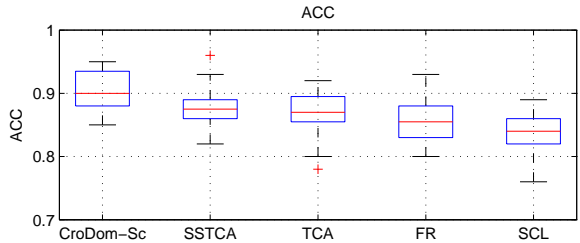}
\caption{The boxplots of detection accuracies of 40 runs for spam detection task.}
\label{fig:FigSpam}
\end{figure}

Figure \ref{fig:FigSpam} shows the boxplots of classification accuracies on the spam detection task.
As we can observed from the figure,
the proposed CroDomSc always outperforms
its competitors.
% over almost all the performance measures.
This is another solid evidence of the effectiveness of
the sparse coding method
for the cross-domain representation problem.
Moreover, SSTCA, which is also a semi-supervised cross-domain representation method, seems to outperform
other methods in some cases. However,
the differences of its performances and other ones are not significant.
%These results again
%confirm the effectiveness of sparse coding for capturing latent
%structure in spam detection task.

\section{Conclusion}
\label{sec:Concl}

In this paper, we introduce the first sparse coding algorithm for cross-domain data representation
problem. The sparse code distribution differences between source and target domains
are reduced by regularizing sparse codes with MMD criterion.
Moreover, the class labels of both source and target domain samples are utilized to encourage the
discriminative ability. The developed cross-domain sparse coding algorithm is tested on
two cross-domain learning tasks and the effectiveness was shown.

\section*{Acknowledgements}
This work was supported by
the
National Key Laboratory for Novel Software Technology, Nanjing University
(Grant No. KFKT2012B17).

\vfill\eject

\balancecolumns
% That's all folks!
\end{document}